\documentclass[pmlr]{jmlr} 

\newcommand{\tagdsmode}{nonproceedings}

\makeatletter
\newcommand{\tagdssubmission}{submission}
\newcommand{\tagdsproceedings}{proceedings}

\ifx\tagdsmode\tagdsproceedings

\else\ifx\tagdsmode\tagdssubmission
  \def\ps@jmlrtps{%
    \let\@mkboth\@gobbletwo
    \def\@oddhead{\scriptsize Under Review at the 2nd Conference on Topology, Algebra, and Geometry in Data Science\hfill}%
    \let\@evenhead\@oddhead
    \def\@oddfoot{}%
    \let\@evenfoot\@oddfoot
  }

\else
  \def\ps@jmlrtps{%
    \let\@mkboth\@gobbletwo
    \def\@oddhead{}%
    \let\@evenhead\@oddhead
    \def\@oddfoot{}%
    \let\@evenfoot\@oddfoot
  }
\fi\fi
\makeatother

\usepackage{enumitem}
\usepackage{graphicx}
\usepackage{booktabs}
\usepackage{tcolorbox}
\tcbuselibrary{listings}
\usepackage{fancyvrb}
\usepackage{xcolor}

\usepackage{longtable}

\usepackage{booktabs}
\usepackage[load-configurations=version-1]{siunitx} 

\theorembodyfont{\upshape}
\theoremheaderfont{\scshape}
\theorempostheader{:}
\theoremsep{\newline}

\jmlrvolume{334}
\jmlryear{2026}
\jmlrworkshop{Topology, Algebra, and Geometry in Data Science}

\newcommand{\jointfirst}{$^*$}

\title[Evaluating Mathematical Equivalence in Embedding Models]{Does My Embedding Reflect That \(A = B\)? Evaluating Mathematical Equivalence in Embedding Models}

\ifx\tagdsmode\tagdssubmission

\else

  \author{\Name{Jiaying Ye\nametag{\thanks{Joint first authors.}}} \Email{yjy0509@uw.edu} \\
   \Name{Samarth Rao\nametag{\jointfirst}} \\
   \Name{Leo Carlin\nametag{\jointfirst}} \\
   \Name{Kedar Chintalapati} \\
   \Name{Saharsh Bhargava} \\
   \Name{Rachit Jaiswal} \\
   \Name{Michael Zhou} \\
   \Name{Jared Darlington} \\
   \Name{Jiahe Lu}\\
   \Name{Jarod Alper}\\
   \addr  University of Washington
   \AND
   \Name{Vasily Ilin} \\
   \addr  Axiom Math, University of Washington
   \AND
   \Name{Henry Kvinge} \Email{hjk3@uw.edu}\\
   \addr University of Washington, Pacific Northwest National Laboratory
   }

\fi

\ifx\tagdsmode\tagdsproceedings
\editor{Editor's name}
\fi

\begin{document}

\maketitle

\begin{abstract}
Because mathematics is highly abstract, a single statement can take very different forms depending on what subfield it is framed in. There are many examples where breakthroughs occurred after researchers discovered that a question had already been answered in a different field. At the same time, the growth of new resources related to formalization has increased the need for tools that enable efficient and reliable navigation between mathematical `languages’ (e.g., from Lean to natural language). In this paper, we investigate whether current embedding models capture mathematical equivalence. To do this, we introduce the \emph{Mathematically Equivalent but Lexically Different Pairs (MELD) Dataset}, a collection of mathematically equivalent statements that are expressed in very different language. We show that current state-of-the-art embedding models tend to group statements by the terminology used to make them instead of the underlying math. Motivated by this, we propose a contrastive approach to learning embeddings of mathematical text that focuses on aligning informal statements with different formalizations. Our experiments demonstrate that this leads to improvements not only on informal-formal retrieval tasks but also on MELD, which only contains natural language statements.
\end{abstract}

\section{Introduction}
\label{sec:tag-ds}

Language models have become increasingly capable of research-level mathematics with several documented cases of LLMs like GPT \citep{tao2024primetime,gowers2026chatgpt,openai2026unitdistance}, Gemini \citep{feng2026semi}, and Claude \citep{knuth2024news} helping mathematicians make progress on open problems. One of the advantages of these systems is their ability to draw on the vast mathematics literature (through both parametric knowledge and internet search) at a scale that is inaccessible to humans. This includes making non-trivial connections that extend beyond surface-level lexical overlap.

At the same time, the ability to effectively navigate large collections of mathematical text is becoming essential not only because of the ever-growing number of mathematics papers available in online databases like the arXiv, but also because of new formalization-driven resources like \texttt{mathlib} \citep{mathlib2020}. Although state-of-the-art LLMs may already have the capability to perform this type of navigation, they are too expensive computationally to scale to many applications. On the other hand, text embedding models are explicitly designed to capture semantic relationships within a corpus of text efficiently. In this paper, we ask whether current state-of-the-art embedding models can capture mathematical equivalence.

To test this, we introduce a new dataset, \emph{Mathematically Equivalent Lexically Different Pairs (MELD)}, a collection of 270 pairs of mathematically equivalent statements expressed in different terminology. One simple example is the definition of a product of two sets in terms of set theory versus category theory. We then show that state-of-the-art embedding models tend not to embed these pairs near one another, instead grouping statements based on the mathematical subfield they are framed in.

An embedding model that could group statements based on mathematical equivalence rather than lexical similarity would be a powerful tool for mathematical discovery. Motivated by this, we propose a contrastive approach to training embedding models for mathematics that uses formal and informal versions of a statement as different ``views'' of the same underlying mathematics. To realize this, we generate an augmented version of the \verb|mathlib_informal_v4.16.0| dataset \citep{frenzymath2025mathlibinformal}, which contains Lean statements from the \texttt{mathlib} library along with informal descriptions. We also add rephrasings of the informal descriptions. The two models we train under this setup, \verb|MathLeap-Qwen-8B| and \verb|MathLeap-Octen-8B|, achieve strong results across a range of mathematics retrieval tasks\footnote{Code can be found at \url{https://github.com/uw-math-ai/math2vec}.}. Supporting our hypothesis that our approach to training encourages a model to learn representations that reflect the mathematical content of a statement, we show that both models achieve the highest scores on MELD from among the models that we evaluated. This is particularly notable since MELD only contains natural language content.

In summary, our contributions include:
\begin{enumerate}[itemsep=0pt, topsep=2pt]
    \item We introduce \emph{Mathematically Equivalent but Lexically Different Pairs (MELD)}, a dataset with 270 pairs of mathematical statements that are lexically dissimilar but mathematically equivalent.
    \item Observing that most embedding models prioritize language similarity over mathematical equivalence, we propose a contrastive approach to developing embedding models for mathematics, using informal and formal representations of a statement as different views of the underlying mathematical content.
    \item We show that this leads to significant improvements on MELD and a range of other mathematical retrieval tasks.
\end{enumerate} 

\section{Related work}

The value of being able to efficiently leverage the mathematics literature has meant that retrieval and search are an active area of research in the AI-for-math community. For example, \citep{welleck2021naturalproofs} introduced a dataset of mathematical theorems, treating theorem search as a reference retrieval task. This showed that text embedding models can be effective when retrieval only depends on capturing surface-level similarities between mathematical statements but may be less effective when deeper mathematical connections are required. More recently, \citep{alexander2026semantic} used embeddings from Qwen3-Embedding-8B to perform retrieval over 9 million theorems. On the other hand, our work aims to improve the capabilities of text embedding models themselves, focusing specifically on capturing mathematical equivalence.

More recently \citep{ju2025mirb}, proposed a full text embedding benchmark for mathematical retrieval, Mathematics Information Retrieval Benchmark (MIRB). This benchmark includes tasks that mix less rigorous language (e.g., search for MathStackExchange duplicates) to more formal language (e.g., Lean premise retrieval). Our work differs from MIRB, as MELD is designed to target a specific embedding model capability rather than general mathematics retrieval.

\section{Do Embedding Models Capture Mathematical Equivalence?}
\label{sect:amp}

\begin{figure}[t]
    \centering
    \includegraphics[width=0.5\columnwidth]{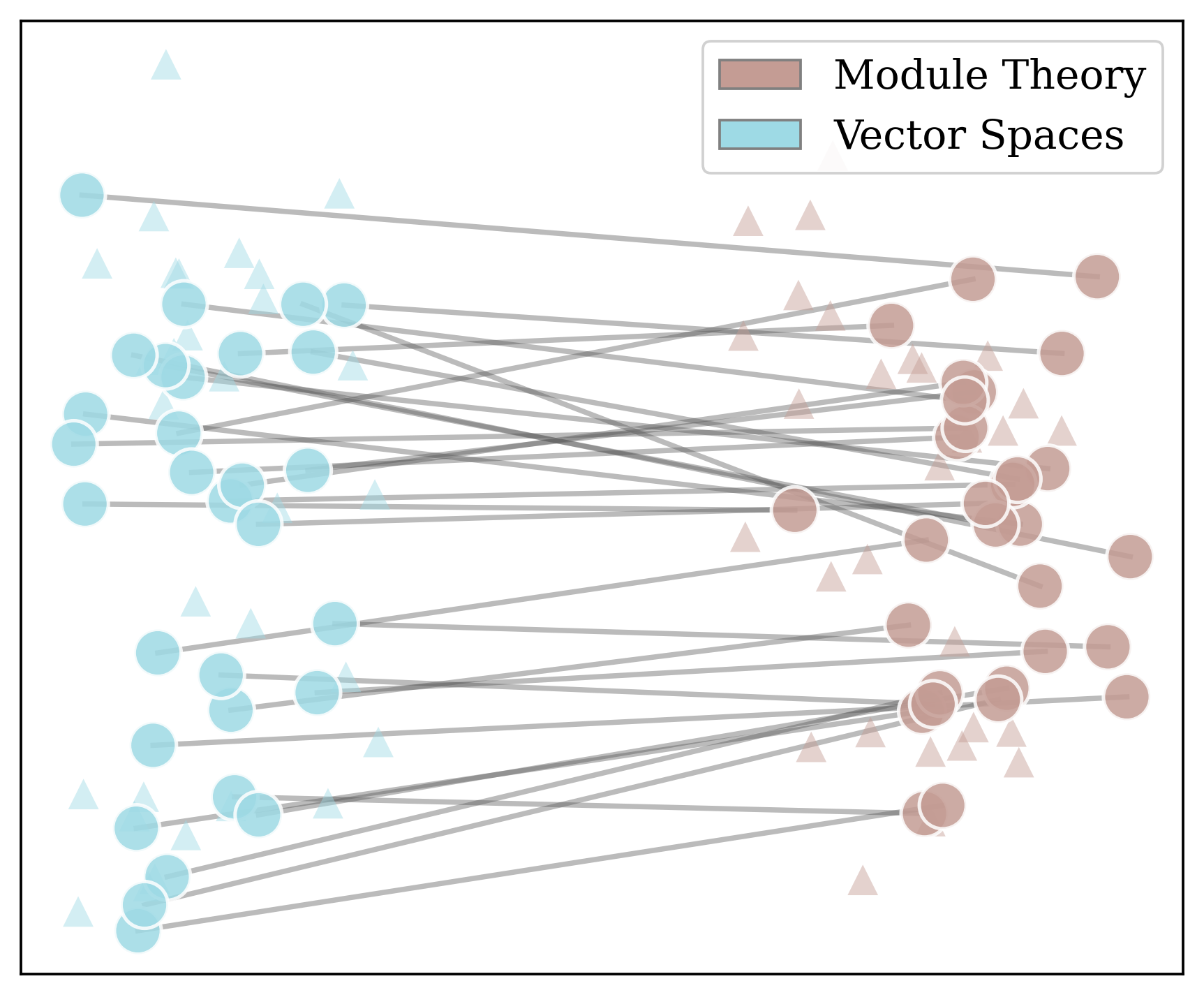}
    \caption{A UMAP visualization of embedded statements from the ``vector spaces vs module theory'' subset of MELD. Blue points correspond to statements framed in terms of vector space terminology, brown points correspond to statements framed in terms of module theory. Small triangles are distractor statements. MELD pairs are large dots connected by lines. Embedded statements cluster by subfield rather than mathematical equivalence.}
    \label{fig:vector-spaces-vs-module-theory}
\end{figure}

One of the most attractive features of mathematics (and also one of the features that can make mathematics so hard) is that a single statement can often be described precisely in two radically different ways. These ``one thing, two perspectives'' situations are the foundation of many fruitful areas of mathematics (e.g., algebraic geometry, algebraic topology, and geometric group theory). 

\begin{table*}[htbp]
\centering
\caption{Subfield pairs used to generate MELD}
\label{tab:adversarial_pair_areas}
\begin{tabular}{lc}
\toprule
Mathematical subfield 1 & Mathematical subfield 2  \\
\midrule
Vector spaces & Module theory \\
Measure theory & Probability theory \\
Set theory & Category theory \\
Geometry & Commutative algebra \\
Algebraic topology & Group theory \\
Graph theory & Linear algebra \\
Combinatorics & Generating functions \\
Representation theory & Fourier analysis \\
Symmetric functions & Tableaux \\
\bottomrule
\end{tabular}
\end{table*}

Motivated by this, we introduce \emph{Mathematically Equivalent but Lexically Different Pairs (MELD)}\footnote{MELD can be found at \url{https://huggingface.co/datasets/uw-math-ai/MELD-dataset}}, a collection of 270 statement pairs written in natural language and LaTeX where paired statements are equivalent but cannot be easily matched by surface-level lexical similarity. One example is the following:
\begin{quote}
\noindent\textbf{Identity map (Set theory)}
For a set $A$, the map $\mathrm{id}_A : A \to A$ defined by $\mathrm{id}_A(a) = a$.
\end{quote}
\begin{quote}
\noindent\textbf{Identity map (Category theory)}
For each object $A$, a distinguished arrow $\mathrm{id}_A : A \to A$ such that $f \circ \mathrm{id}_A = f$ and $\mathrm{id}_A \circ g = g$ for any objects $B$ and $C$ and $f:A \rightarrow B$ and $g: C \rightarrow A$.
\end{quote}
The 270 pairs come from nine different pairs of domains that offer distinctive descriptions of a common set of ideas (e.g., representation theory and Fourier analysis). These are summarized in Table \ref{tab:adversarial_pair_areas}.

\begin{table*}[htbp]
\centering
\small
\caption{Retrieval recall of mathematically equivalent statements from MELD. The best score or scores for each metric are written in bold.}
\label{tab:amp_recall_specialized}
\begin{tabular}{lccccc}
\toprule
Model & Recall@1 & Recall@3 & Recall@5 & Recall@10 & Recall@20 \\
\midrule
Qwen3-Embedding-4B & 13.7 & 34.3 & 45.0 & 58.5 & 70.2 \\
Qwen3-Embedding-8B & 17.0 & 38.0 & 47.8 & 63.0 & 78.1 \\
harrier-oss-v1-27b & 19.4 & 37.0 & 48.3 & 63.9 & 78.5 \\
KaLM-Embedding-Gemma3-12B & 10.4 & 27.4 & 35.2 & 48.9 & 64.6 \\
llama-embed-nemotron-8b & 2.8 & 7.2 & 11.1 & 19.3 & 31.7 \\
Octen-Embedding-8B & 25.0 & 50.6 & 62.8 & 77.4 & 88.9 \\
MathLeap-Qwen-8B (ours) & 27.2 & 52.2 & 63.1 & \textbf{78.0} & \textbf{89.8} \\
MathLeap-Octen-8B (ours) & \textbf{28.9} & \textbf{55.0} & \textbf{63.7} & 77.2 & \textbf{89.8} \\
\bottomrule
\end{tabular}
\end{table*}

We generated MELD by iterating through each of the nine pairs of complementary domains, describing the connection between the two fields, and prompting Claude Opus 4.7 to generate 30 pairs of mathematically equivalent but lexically distinct statements. We then manually reviewed these, making modifications to increase dissimilarity while preserving equivalence. We then evaluated all statements using GPT-5.5 (medium) to check (i) whether both statements were valid, (ii) whether they were equivalent, and (iii) whether they could be made to sound less similar. 

While using LLM-generated datasets carries risks, all content belongs to a basic graduate curriculum in mathematics, and hence should be well within the capabilities of current LLMs. This also means that embedding model failures on MELD are less likely to be attributable to the subject matter being esoteric and more likely to point toward limitations in the way that these models represent mathematical text.

To understand the extent to which embeddings of mathematical text are organized by mathematical equivalence rather than just lexical similarity, we embed all 540 statements using some of the best current embedding models and look at whether equivalent pairs are grouped near each other. This includes \verb|Qwen3-Embedding-4B| \citep{qwen3embedding}, \verb|Qwen3-Embedding-8B| \citep{qwen3embedding}, \verb|harrier-oss-v1-27b| \citep{microsoft2026harrier}, \verb|KaLM-Embedding-Gemma3-12B| \citep{hu2025kalmembedding,zhao2025kalmembeddingv2}, \verb|llama-embed-nemotron-8b| \citep{babakhin2025llamaembednemotron8buniversaltextembedding}, and \verb|Octen-Embedding-8B| \citep{octen2025rteb}. 

We find that statements are mostly clustered by the field they are expressed in. For instance, in Table \ref{tab:amp_recall_specialized} the model that achieves the best results, Octen-Embedding-8B, has a recall@1 of only 25.0. To get a qualitative sense of why this is, in Figure \ref{fig:vector-spaces-vs-module-theory} we use UMAP \citep{mcinnes2018umap} to reduce the dimensionality of the embedding and then visualize statements from 30 pairs where one element is written in the language of vector spaces (blue) and the other is written in the language of module theory (brown). Paired statements are connected by gray lines. The model appears to strongly clusters statements by field rather than equivalence.

\section{A Contrastive Approach to Embedding Mathematical Text}

The experiments in Section \ref{sect:amp} suggest that current embedding models capture the lexical similarity of mathematical statements rather than their mathematical equivalence. Motivated by this, we consider training strategies that steer models toward embeddings that cluster statements by mathematical equivalence rather than surface-level similarities in language. While training on MELD has the potential to do this, it is small and there is not an easy way to cheaply scale it since it depends on state-of-the-art LLMs. Fortunately, the growth of formalization has begun to create new resources that address this need.

\texttt{mathlib} is essential for formal proof development in Lean because it provides formalizations of central definitions and theorems from modern mathematics. \texttt{mathlib} was used to generate the \verb|mathlib_informal_v4.16.0| dataset \citep{frenzymath2025mathlibinformal}. Each entry in this dataset corresponds to a \texttt{mathlib} statement, which comes in three different forms: the Lean signature, the Lean type, and a natural language summary. We treat these as different ``views'' on a single underlying mathematical statement. Our assumption is that a consequence of steering a model toward alignment of formal and informal versions of statements, will be embeddings that better capture mathematical equivalence. 

\subsection{Training Our MathLeap Models}

In addition to the natural language, Lean signature, and Lean type views inherited from \verb|mathlib_informal_v4.16.0|, we also generate natural language rephrasings for some instances in the training set. We do this by first using \verb|Qwen3.5-9B| to decompose natural language theorems into explicit hypotheses and conclusions. We then provide these to \verb|gemma-4-E4B-it| to generate rephrasings. When doing this, \verb|gemma-4-E4B-it| is prompted to change superficial aspects of the theorem statement like variable names. More details, including the prompts that were used, can be found in Section \ref{sect:training-data-generation}. Our final training set contains 118,334 instances (with three to four views each) while our test set contains 15,287 instances\footnote{\url{https://huggingface.co/datasets/uw-math-ai/Math2Vec-embedding-dataset}}. Approximately 85\% of the training set had a natural language rephrasing attached to it.

Since our initial experiments showed that \verb|Qwen3-Embedding-8B| \citep{qwen3embedding} and \verb|Octen-Embedding-8B| \citep{octen2025rteb} are the most effective existing models for mathematical retrieval tasks, we chose to use these as base models for finetuning. We trained using Cached Multiple Negatives Ranking Loss from the Sentence-BERT library \citep{reimers-2019-sentence-bert}, a contrastive loss function that uses in-batch negatives while avoiding common issues that arise when trying to scale contrastive methods to GPU training. We used the default scale value of 20.0. We used AdamW, 700 warm-up steps, a constant scheduler, a batch size of 16, and a max sequence length of 128. Models were trained for approximately two epochs ($\sim$14,790 steps). Training was performed on a single H200 and took approximately 8 hours. We call the resulting models \verb|MathLeap-Qwen-8B|\footnote{\url{https://huggingface.co/uw-math-ai/MathLeap-Qwen-8B}} and \verb|MathLeap-Octen-8B|\footnote{\url{https://huggingface.co/uw-math-ai/MathLeap-Octen-8B}} respectively. 

\subsection{Evaluation}

We evaluate \verb|MathLeap-Qwen-8B| and \verb|MathLeap-Octen-8B| on a range of mathematical retrieval tasks. Overall, we find that our contrastive approach to training a mathematics embedding model significantly improves performance on tasks where retrieving mathematically equivalent statements is critical.

\paragraph{MELD} Finetuning \verb|Qwen3-Embedding-8B| results in recall@1 increasing from 17.0 to 27.2 and recall@5 increasing from 47.8 to 63.1 (Table \ref{tab:amp_recall_specialized}). At the same time, the average rank of the mathematically equivalent pair among all neighbors decreases from 16.1 to 9.7 (Table \ref{tab:amp_ranking}). Both \verb|MathLeap-Qwen-8B| and \verb|MathLeap-Octen-8B| independently score higher on MELD than any of the other models we evaluated on each of the metrics we considered. This is particularly notable since natural-language-only retrieval is a small part of the training curriculum of our models. We see this as supporting our hypothesis that our contrastive style of training does not just help with the specific task of aligning representations of formal and informal mathematical statements, it also encourages representations that capture mathematical equivalence.

\begin{table*}[htbp]
\centering
\small
\caption{Retrieval scores on the blueprints dataset. Retrieval was performed by giving the informal description and asking the model to find the corresponding Lean statement. The best score or scores for each metric are written in bold.}
\label{tab:amp_ranking}
\begin{tabular}{lccc}
\toprule
Model & Recall@1 & Recall@5 & Recall@10 \\
\midrule
harrier-oss-v1-27B & 55.00 & 85.82 & 92.07\\
Qwen3-Embedding-4B & 56.94 & 88.76 & 94.11 \\
Qwen3-Embedding-8B & 59.52 & 90.13 & 94.11 \\
KaLM-Embedding-Gemma3-12B-2511 & 54.04 & 86.02 & 91.96 \\
llama-embed-nemotron-8b & 23.80 & 44.11 & 52.70 \\
Octen-Embedding-8B & 64.32 & 92.60 & 96.37 \\
MathLeap-Qwen-8B (ours) & \textbf{64.73} & \textbf{93.10} & 96.44  \\
MathLeap-Octen-8B (ours) & 64.69 & 92.80 & \textbf{96.92} \\
\bottomrule
\end{tabular}
\end{table*}

\paragraph{Blueprints} Lean blueprints are dependency graphs used to track progress on formalization projects. Nodes in these graphs typically contain both a Lean statement and a natural language description. We used these pairs across 68 graphs to create a dataset of 3,729 Lean and natural language pairs. We measured recall on retrieval of the corresponding Lean statement given a natural language query. Across all models that we evaluated, \verb|MathLeap-Qwen-8B| or \verb|MathLeap-Octen-8B| performed best across the evaluated recall metrics (Table \ref{tab:amp_ranking}). This is not as surprising as our results on MELD, since the blueprints task aligns closely with the training task for both MathLeap models.

\begin{table*}[htbp]
\centering
\small
\caption{nDCG@10 scores for three MIRB tasks. The best score or scores for each metric are written in bold.}
\label{tab:amp_ranking}
\begin{tabular}{lccc}
\toprule
Model & MO Dup. & \texttt{mathlib} Retrieval & Lean Premise \\
\midrule
harrier-oss-v1-0.6B & 0.762 & 0.345 & 0.087\\
Qwen3-Embedding-4B & 0.855 & 0.531 & 0.132 \\
Qwen3-Embedding-8B & \textbf{0.859} & 0.559 & 0.136 \\
KaLM-Embedding-Gemma3-12B-2511 & 0.778 & 0.504 & 0.133 \\
llama-embed-nemotron-8b & 0.730 & 0.267 & 0.050 \\
Octen-Embedding-8B & 0.855 & 0.653 & \textbf{0.143} \\
MathLeap-Qwen-8B (ours) & 0.821 & 0.653 & 0.103  \\
MathLeap-Octen-8B (ours) & 0.816 & \textbf{0.667} & 0.103 \\
\bottomrule
\end{tabular}
\end{table*}

\paragraph{MIRB} We evaluate \verb|MathLeap-Qwen-8B| and \verb|MathLeap-Octen-8B| on three tasks from the MIRB benchmark: MathOverflow duplicate question retrieval, \texttt{mathlib} retrieval, and Lean premise retrieval (originally from \citep{yang2023leandojo}). We measure performance using nDCG@10, following \citep{ju2025mirb}. \verb|MathLeap-Octen-8B| performs best on \texttt{mathlib} retrieval, which is well-aligned with its training set. On the other hand, we see reduced performance on the other two tasks. We analyze these trade-offs in the Limitations section (Section \ref{sect:limitations}).

\section{Limitations}
\label{sect:limitations}

There are a number of limitations to this work that should be highlighted. Most importantly, we note that on some tasks such as MathOverflow duplication retrieval and Lean premise retrieval, the performance of \verb|MathLeap-Qwen-8B| and \verb|MathLeap-Octen-8B| decreases relative to their base models. We suspect this is due to our models becoming over-specialized during finetuning. For example, MathOverflow duplication retrieval may require a broader natural-language understanding outside of rigorous theorem statements, an ability that may degrade in our models during finetuning. On the other hand, Lean premise retrieval involves using a proof-state as a query and \texttt{mathlib} declarations as the corpus. Our training set does not include any proof-states so it is not surprising that performance degrades. Overall, this points to the need for a more diverse training set if our goal is a generalist math embedding model.

We also note that we depend on synthetic data for a number of our datasets. MELD depends on frontier LLMs for identification and generation of basic theorems and definitions, as well as translation of these to the language of a different mathematical subfield. The training set of \verb|MathLeap-Qwen-8B| and \verb|MathLeap-Octen-8B| uses rephrasing generated by local models. In both cases, we have attempted to remove any erroneous instances, but we cannot guarantee that all errors have been eliminated.

\section{Conclusion}
\label{sec:conclusion}

The ability of an embedding model to capture mathematical equivalence rather than just surface-level textual similarity is important to mathematical discovery. In this paper, we ask whether today's state-of-the-art embedding models are capable of retrieving mathematically equivalent statements from among lexically similar distractors. Using a new dataset, Mathematically Equivalent but Lexically Different Pairs (MELD), we find that current models struggle to do this. This leads us to propose a contrastive approach to mathematics-specific embedding model training using informal and formal statements as different views of the underlying mathematics. Using this approach, we train \verb|MathLeap-Qwen-8B| and \verb|MathLeap-Octen-8B|. These models display substantial improvement across many mathematics retrieval tasks and suggest a path toward training even stronger mathematics specific embedding models in the future.


\bibliography{main}

\appendix

\begin{table*}[htbp]
\centering
\small
\caption{MMR and mean pair rank for embedding models on MELD. The best score or scores for each metric are written in bold.}
\label{tab:amp_ranking}
\begin{tabular}{lcc}
\toprule
Model & MMR & Mean rank \\
\midrule
Qwen3-Embedding-4B & 0.28 & 21.4 \\
Qwen3-Embedding-8B & 0.32 & 16.1 \\
harrier-oss-v1-27b & 0.33 & 15.2 \\
KaLM-Embedding-Gemma3-12B-2511 & 0.23 & 25.6 \\
llama-embed-nemotron-8b & 0.08 & 94.7 \\
Octen-Embedding-8B & 0.42 & 10.3 \\
F2LLM-v2-14B & 0.37 & 10.17  \\
MathLeap-Qwen-8B (ours) & 0.43 & 9.7 \\
MathLeap-Octen-8B (ours) & \textbf{0.45} & \textbf{9.3} \\
\bottomrule
\end{tabular}
\end{table*}

\section{FrenzyMath Retrieval Training and Evaluation}
\label{sec:frenzymath-retrieval-details}

In this section we describe the training and testing details on our augmented version of \texttt{FrenzyMath/mathlib\_informal\_v4.19.0} \cite{frenzymath2025mathlibinformal419}. We focus on mathematics-related instances (as opposed to those related to Lean infrastructure), retaining examples whose \texttt{module\_name} field has a second component in the following set: 
\begin{itemize}
\item \texttt{Algebra}, 
\item \texttt{AlgebraicGeometry}, 
\item \texttt{AlgebraicTopology}, 
\item \texttt{Analysis}, 
\item \texttt{CategoryTheory}, 
\item \texttt{Combinatorics}, 
\item \texttt{Computability}, 
\item \texttt{Condensed}, 
\item \texttt{Control}, 
\item \texttt{Data}, 
\item \texttt{Dynamics}, 
\item \texttt{FieldTheory}, 
\item \texttt{Geometry}, 
\item \texttt{GroupTheory}, 
\item \texttt{InformationTheory}, 
\item \texttt{LinearAlgebra}, 
\item \texttt{Logic}, 
\item \texttt{MeasureTheory}, 
\item \texttt{ModelTheory}, 
\item \texttt{NumberTheory}, 
\item \texttt{Order}, 
\item \texttt{Probability}, 
\item \texttt{RepresentationTheory}, 
\item \texttt{RingTheory}, 
\item \texttt{SetTheory}, 
\item \texttt{Topology}. 
\end{itemize}

Our models are trained on an exact paired-retrieval task. During evaluation, queries are drawn from the held-out test split but the retrieval corpus is drawn from the union of the train, validation, and test splits. The model never sees validation or test examples during training. A retrieved item is counted as correct only if it matches the row-aligned target for the query. All embeddings are $L_2$-normalized, and retrieval is performed by exact nearest-neighbor search in the shared embedding space.

Tables~\ref{tab:frenzymath-informal-type}-\ref{tab:frenzymath-signature-type} report six directed retrieval tasks: informal description to Lean type, Lean type to informal description, informal description to Lean signature, Lean signature to informal description, Lean type to Lean signature, and Lean signature to Lean type. For instruction-aware models, the task instruction is applied only to the query side, while the corpus side is embedded as the fixed candidate bank. We report Recall@1 (R@1), Recall@5 (R@5), Recall@10 (R@10), and mean reciprocal rank (MRR) for the base embedding models used in the main experiments and our fine-tuned variants, \texttt{MathLeap-Qwen-8B} and \texttt{MathLeap-Octen-8B}.



\begin{table*}[htbp]
\centering
\small
\caption{Informal description to Lean type. The best score or scores for each metric are written in bold.}
\label{tab:frenzymath-informal-type}
\begin{tabular}{@{}lrrrr@{}}
\toprule
Model & R@1 & R@5 & R@10 & MRR \\
\midrule

Octen-Embedding-8B & 64.06\% & 87.91\% & 91.43\% & 74.47\% \\
Qwen3-Embedding-8B & 61.89\% & 86.70\% & 90.71\% & 72.65\% \\
Qwen3-Embedding-4B & 57.01\% & 84.33\% & 89.08\% & 68.79\% \\
F2LLM-v2-14B & 67.97\% & \textbf{90.42\%} & \textbf{93.36\%} & 77.78\% \\
Harrier-OSS-v1-27B & 62.26\% & 87.90\% & 91.67\% & 73.37\% \\
Llama-Embed-Nemotron-8B & 27.87\% & 52.79\% & 61.72\% & 38.42\% \\
KaLM-Gemma3-12B & 56.32\% & 82.25\% & 87.49\% & 67.54\% \\
MathLeap-Qwen-8B (ours) & 69.34\% & 88.35\% & 91.72\% & 77.77\% \\
MathLeap-Octen-8B (ours) & \textbf{70.18\%} & 88.27\% & 91.64\% & \textbf{78.26\%} \\
\bottomrule
\end{tabular}%
\end{table*}

\begin{table*}[htbp]
\centering
\small
\caption{Lean type to informal description. The best score or scores for each metric are written in bold.}
\label{tab:frenzymath-type-informal}
\begin{tabular}{@{}lrrrr@{}}
\toprule
Model & R@1 & R@5 & R@10 & MRR \\
\midrule
Octen-Embedding-8B & 58.17\% & 85.56\% & 90.07\% & 69.90\% \\
Qwen3-Embedding-8B & 44.97\% & 74.90\% & 82.61\% & 57.69\% \\
Qwen3-Embedding-4B & 45.33\% & 75.00\% & 82.40\% & 57.98\% \\
F2LLM-v2-14B & 56.78\% & 85.02\% & \textbf{90.09\%} & 68.93\% \\
Harrier-OSS-v1-27B & 33.03\% & 61.87\% & 71.53\% & 45.33\% \\
Llama-Embed-Nemotron-8B & 1.17\% & 3.56\% & 5.78\% & 2.30\% \\
KaLM-Gemma3-12B & 42.31\% & 71.30\% & 79.43\% & 54.69\% \\
MathLeap-Qwen-8B (ours) & 65.22\% & 85.77\% & 89.29\% & 74.26\% \\
MathLeap-Octen-8B (ours) & \textbf{66.89\%} & \textbf{86.54\%} & 89.92\% & \textbf{75.57\%} \\
\bottomrule
\end{tabular}%
\end{table*}

\begin{table*}[htbp]
\centering
\small
\caption{Informal description to Lean signature. The best score or scores for each metric are written in bold.}
\label{tab:frenzymath-informal-signature}
\begin{tabular}{@{}lrrrr@{}}
\toprule
Model & R@1 & R@5 & R@10 & MRR \\
\midrule
Octen-Embedding-8B & 56.08\% & 79.47\% & 84.27\% & 66.28\% \\
Qwen3-Embedding-8B & 54.64\% & 77.95\% & 82.90\% & 64.79\% \\
Qwen3-Embedding-4B & 49.14\% & 75.22\% & 81.18\% & 60.37\% \\
F2LLM-v2-14B & 57.73\% & 81.16\% & 85.94\% & 67.98\% \\
Harrier-OSS-v1-27B & 54.16\% & 78.79\% & 84.04\% & 64.76\% \\
Llama-Embed-Nemotron-8B & 21.89\% & 39.42\% & 46.20\% & 29.47\% \\
KaLM-Gemma3-12B & 47.56\% & 72.83\% & 79.07\% & 58.31\% \\
MathLeap-Qwen-8B (ours) & \textbf{67.40\%} & \textbf{88.75\%} & \textbf{91.96\%} & \textbf{76.64\%} \\
MathLeap-Octen-8B (ours) & 67.24\% & 87.82\% & 91.51\% & 76.28\% \\
\bottomrule
\end{tabular}%
\end{table*}

\begin{table*}[htbp]
\centering
\small
\caption{Lean signature to informal description. The best score or scores for each metric are written in bold.}
\label{tab:frenzymath-signature-informal}
\begin{tabular}{@{}lrrrr@{}}
\toprule
Model & R@1 & R@5 & R@10 & MRR \\
\midrule
Octen-Embedding-8B & 57.44\% & 82.94\% & 87.94\% & 68.48\% \\
Qwen3-Embedding-8B & 42.39\% & 70.67\% & 78.16\% & 54.50\% \\
Qwen3-Embedding-4B & 47.10\% & 73.93\% & 80.39\% & 58.65\% \\
F2LLM-v2-14B & 57.51\% & 84.36\% & 89.47\% & 69.10\% \\
Harrier-OSS-v1-27B & 38.47\% & 66.16\% & 74.51\% & 50.29\% \\
Llama-Embed-Nemotron-8B & 2.62\% & 6.73\% & 9.66\% & 4.43\% \\
KaLM-Gemma3-12B & 43.36\% & 70.78\% & 78.09\% & 55.12\% \\
MathLeap-Qwen-8B (ours) & 71.28\% & 91.98\% & 94.45\% & 80.33\% \\
MathLeap-Octen-8B (ours) & \textbf{72.77\%} & \textbf{92.33\%} & \textbf{94.77\%} & \textbf{81.37\%} \\
\bottomrule
\end{tabular}%
\end{table*}

\begin{table*}[htbp]
\centering
\small
\caption{Lean type to Lean signature. The best score or scores for each metric are written in bold.}
\label{tab:frenzymath-type-signature}
\begin{tabular}{@{}lrrrr@{}}
\toprule
Model & R@1 & R@5 & R@10 & MRR \\
\midrule
Octen-Embedding-8B & 60.40\% & 81.99\% & 86.60\% & 69.81\% \\
Qwen3-Embedding-8B & 58.00\% & 80.41\% & 85.08\% & 67.75\% \\
Qwen3-Embedding-4B & 51.99\% & 77.28\% & 83.13\% & 62.90\% \\
F2LLM-v2-14B & 57.96\% & 81.90\% & 86.82\% & 68.33\% \\
Harrier-OSS-v1-27B & 48.36\% & 72.91\% & 79.04\% & 58.97\% \\
Llama-Embed-Nemotron-8B & 8.81\% & 15.90\% & 19.02\% & 11.84\% \\
KaLM-Gemma3-12B & 45.43\% & 68.88\% & 75.45\% & 55.48\% \\
MathLeap-Qwen-8B (ours) & 65.01\% & 83.83\% & 87.72\% & 73.24\% \\
MathLeap-Octen-8B (ours) & \textbf{65.26\%} & \textbf{84.29\%} & \textbf{88.22\%} & \textbf{73.61\%} \\
\bottomrule
\end{tabular}%
\end{table*}

\begin{table*}[htbp]
\centering
\small
\caption{Lean signature to Lean type. The best score or scores for each metric are written in bold.}
\label{tab:frenzymath-signature-type}
\begin{tabular}{@{}lrrrr@{}}
\toprule
Model & R@1 & R@5 & R@10 & MRR \\
\midrule
Octen-Embedding-8B & 70.04\% & 89.27\% & 92.31\% & 78.45\% \\
Qwen3-Embedding-8B & 69.08\% & 89.10\% & 92.50\% & 77.88\% \\
Qwen3-Embedding-4B & 64.41\% & 86.24\% & 90.19\% & 73.98\% \\
F2LLM-v2-14B & 70.43\% & \textbf{90.41\%} & \textbf{93.63\%} & 79.22\% \\
Harrier-OSS-v1-27B & 61.74\% & 84.45\% & 88.69\% & 71.52\% \\
Llama-Embed-Nemotron-8B & 16.57\% & 28.67\% & 33.59\% & 21.79\% \\
KaLM-Gemma3-12B & 67.16\% & 87.33\% & 90.83\% & 76.03\% \\
MathLeap-Qwen-8B (ours) & 72.46\% & 89.01\% & 92.06\% & 79.82\% \\
MathLeap-Octen-8B (ours) & \textbf{73.34\%} & 89.68\% & 92.59\% & \textbf{80.56\%} \\
\bottomrule
\end{tabular}%
\end{table*}

\section{Training data generation}

\subsection{Prompts for theorem decomposition, rephrasing, and hard negative generation}
\label{sect:training-data-generation}

In this section we provide the prompts that we used to generate the natural language rephrasings and hard negatives. The first step in this process was to down select to theorems in \verb|FrenzyMath/mathlib_informal_v4.16.0|. We then had \verb|Qwen/Qwen3.5-9B| decompose each of these into assumptions and conclusions using the following prompt:
\begin{Verbatim}
You are a mathematical statement parser.

Task:
Given a single research-level 
mathematical statement
(theorem, lemma, proposition, 
corollary, claim, definition-like 
implication, or assertion), decompose it 
into:

1. hypotheses/assumptions
2. conclusions
Also rewrite the theorem in 
normalized_form which must 
be a single string of the form:
  "If [H1] and [H2] and ... then [C]."
\end{Verbatim}
The idea is that by explicitly stating the assumptions and conclusion, an LLM will be less likely to change the theorem in ways that make it mathematically different when rephrasing and similarly, it will be easier to control how an LLM changes a theorem to make it a hard negative (we generated hard negatives, as we describe below, but these were not used in development of the final models).

We gave theorem decompositions to \verb|google/gemma-4-E4B-it| and requested rephrasings using the prompt:
\begin{Verbatim}
You are a mathematical writer helping 
to build a training dataset for an \
embedding model. Your task is to 
produce alternative natural language \
phrasings of mathematical statements 
that preserve their meaning exactly.

You are given the original statement 
together with a pre-computed 
decomposition into hypotheses and 
conclusions. Use that decomposition 
as your guide to reassemble the 
components into a rephrasing with 
a different surface form.

The rephrasing must be mathematically 
identical to the original — it must \
make the same claims with the same 
assumptions — but it should differ 
in at least one of the following ways:
  - Sentence structure (e.g. "If A 
  then B" → "Whenever A, we have B")
  - Logical connective style ("Let ... 
  Then ..." → "Given ..., it follows ...")
  - Quantifier ordering or grouping
  - Passive vs active voice
  - Mathematical phrasing conventions 
  (e.g. "there exists" → "we can find")
  - Naming convention (e.g. "a 
  subgroup H" → "a subgroup, which we 
  call H")
  - Consistently swap variables so 
  "a + b = c" becomes "s + r = p" but 
  be sure to keep conistency through 
  the whole statement. 
Overall, the phasing should be as 
different as possible while remaining
completely mathematically equivalent. 

The rephrasing should read as natural 
mathematical prose — the kind you would \
find in a textbook or research paper. 
Include LaTeX notation where the \
original does.

Output the rephrased statement 
directly — no preamble, no explanation, \
no markup. Just the rephrasing itself.
"""

USER_TMPL = """\
Mathematical statement:
\"\"\"STATEMENT\"\"\"

Pre-computed decomposition:
Hypotheses:
HYPOTHESES
Conclusions:
CONCLUSIONS

Rephrasing:"""

RETRY_SUFFIX = (
    "\n\nYour previous response 
    was empty. "
    "Output only the rephrased 
    statement — no preamble 
    or explanation."
)
\end{Verbatim}
Finally, we also prompted \verb|Qwen/Qwen3.5-9B| to give us three hard negatives for the theorem using the prompt (note that we found that training with hard negatives degraded performance so they were omitted from the final training routine):
\begin{Verbatim}
You are an expert mathematical editor.

Task:
Given a string with 
"input_statement", "hypotheses", 
"conclusions", "normalized_form", 
"subject"), generate exactly 3 
"hard_negatives" — statements 
that closely mimic the visual and 
syntactic structure of the 
"input_statement" but are strictly 
mathematically false or 
semantically different.

Guidelines for Hard Negatives 
(This is not an exhaustive list 
of guidelines):
1. Subtle Omission: If it would 
create a mathematically different 
statement then omit a property that 
is a prerequisite for the conclusion, 
or remove a specific conclusion 
2. Commutative Flip: If it would 
create a mathematically different 
statement then reverse the order of 
a product that is not commutable in 
a way that is plausible but strictly 
breaks the equality.
3. Negation: If a negation would 
create a mathematically different 
statement, then negate an assumption 
or conclusion to fundamentally alter 
the semantic meaning of the statement.

Validity: 
- Every variable used in the conclusion 
must be defined in the hypotheses. 
Every operator must act on the correct 
type of object.
- The statement must be read as a 
valid, grammatically correct mathematical 
sentence, even though the underlying 
claim is false.
- It cannot simply be a "weaker" true 
statement. If the original theorem 
guarantees x > 0, generating a negative 
that says x \ge 0 is invalid because it 
is still a true mathematical statement. 
Do not merely omit a conclusion or 
weaken an inequality without introducing 
a strict contradiction or a false 
mathematical guarantee.

Style Constraints:
- Preserve the original LaTeX formatting 
style, notation, and complexity.
- The visual "weight" of each equation 
must match the original theorem exactly.
\end{Verbatim}

\subsection{The role of hard negatives}

We generated hard negatives for informal text that were designed to look lexically similar but have a different mathematical meaning. Often, this modification made the statement false.

During initial fine-tuning runs, we attempted to use the hard negatives alongside the rephrasings. However, we found that with our training techniques, fine-tuning with the hard negatives degraded performance. We therefore decided to only use the rephrasings. We leave hard negatives as a tool to explore in future research.

\end{document}